\documentclass[sigconf]{acmart}
\usepackage{makecell}  
\usepackage{multirow}
\usepackage{booktabs}
\usepackage{makecell}
\AtBeginDocument{%
  }

\setcopyright{acmlicensed}
\copyrightyear{2025}
\acmYear{2025}
\acmDOI{XXXXXXX.XXXXXXX}
\acmConference[CIKM '25]{The 34th ACM International Conference on Information and Knowledge Management}{November 10--14, 2025}{Seoul, Korea}

\acmISBN{978-1-4503-XXXX-X/2018/06}

\begin{document}

\title{\texttt{TalkDep}: Clinically Grounded LLM Personas for Conversation-Centric Depression Screening}

\author{Xi Wang$^*$}
\email{xi.wang@sheffield.ac.uk}
\affiliation{%
  \institution{University of Sheffield}
  \city{Sheffield}
  \country{United Kingdom}
}

\author{Anxo Perez$^*$}
\affiliation{%
  \institution{University of A Coruña}
  \city{A Coruña}
  \country{Spain}}
\email{anxo.pvila@udc.es}

\author{Javier Parapar}
\affiliation{%
  \institution{University of A Coruña}
  \city{A Coruña}
  \country{Spain}}
\email{javier.parapar@udc.es}

\author{Fabio Crestani}
\affiliation{%
 \institution{Università della Svizzera italiana}
 \city{Lugano}
 \country{Switzerland}}
\email{fabio.crestani@usi.ch}

\renewcommand{\shortauthors}{Xi et al.}

\begin{abstract}
The increasing demand for mental health services has outpaced the availability of real training data to develop clinical professionals, leading to limited support for the diagnosis of depression. This shortage has motivated the development of simulated or virtual patients to assist in training and evaluation, but existing approaches often fail to generate clinically valid, natural, and diverse symptom presentations. In this work, we embrace the recent advanced language models as the backbone and propose a novel clinician-in-the-loop patient simulation pipeline, \texttt{TalkDep}, with access to diversified patient profiles to develop simulated patients. By conditioning the model on psychiatric diagnostic criteria, symptom severity scales, and contextual factors, our goal is to create authentic patient responses that can better support diagnostic model training and evaluation. We verify the reliability of these simulated patients with thorough assessments conducted by clinical professionals. The availability of validated simulated patients offers a scalable and adaptable resource for improving the robustness and generalisability of automatic depression diagnosis systems. 
\end{abstract}



\maketitle
\def\thefootnote{*}\footnotetext{Xi and Anxo contributed equally to this work}\def\thefootnote{\arabic{footnote}}
\begin{figure*}
    \centering
    \includegraphics[width=0.8\linewidth]{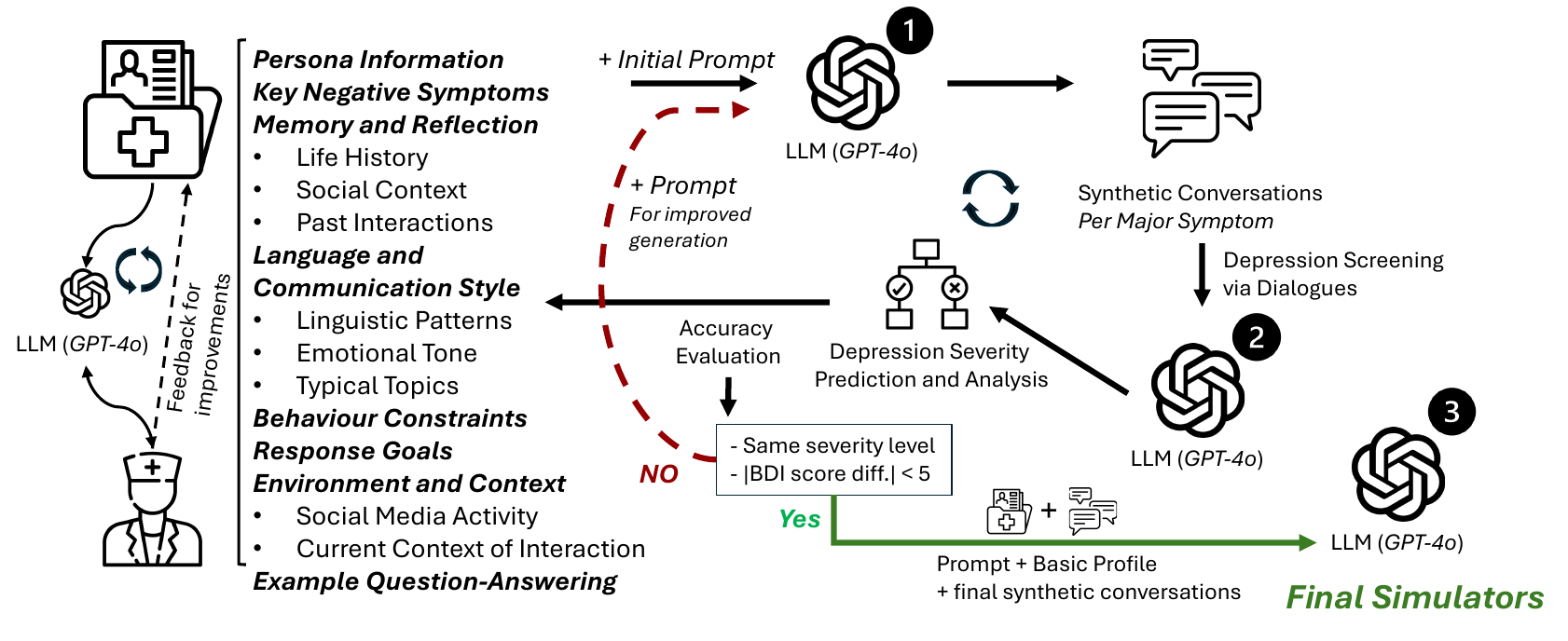}
    \caption{Overview of \texttt{TalkDep}, a patient simulation pipeline.}
    \label{fig:overview}
\end{figure*}

\section{Introduction}

Major depressive disorder (MDD) affects hundreds of millions of people worldwide and is now the single largest contributor to the number of years lived with disability~\cite{lu2024global}. Global estimates put the number of people living with depression at well over 300 million, yet most of them do not receive adequate care, especially in low- and middle-income countries with merely 8\% coverage of mental health treatment~\cite{moitra2022global}. The gap is compounded by the lack of qualified mental health professionals~\cite{lawrence2024opportunities}. This mismatch between demand and clinical capacity has led to intense research into depression screening systems that can operate at scale and in real time.

Since 2017, the CLEF eRisk lab has provided a workshop on early risk detection tasks based on social media publications. In 2025, eRisk introduced a new pilot track \href{https://erisk.irlab.org}{\emph{Conversational Depression Detection via LLM Personas}}, which shifts evaluation to an interactive, dialogue-based setting: participants must engage with LLM personas and decide, as early as possible, whether each persona shows signs of depression, the severity level of depression and major symptoms~\cite{parapar2025erisk,erisk2024,erisk2023,erisk2022}. Although the task opens exciting avenues for studying conversation-centric screening, it also highlights a crucial bottleneck: the field lacks a publicly available, clinically grounded collection of simulated patients that (i) cover clinical guidelines such as the BDI-II~\cite{dozois1998psychometric}, (ii) respond coherently across diverse life contexts, and (iii) are supervised by professionals.

We address this gap by introducing \texttt{TalkDep}, a \emph{language model-driven} and \emph{clinician-in-the-loop} patient simulation pipeline that couples structured depression profiles with free-form generative text.  Each persona of the patient is instructed with (a) explicit demographic and biographical attributes, (b) BDI-II-aligned symptom severities, (c) dynamic context templates that guide conversational behaviour, and (d) conversations exemplify expression styles.  A set of consistency and safety checks -- both automatic and human -- filter responses to ensure clinical plausibility and ethical compliance.  The resulting resource currently comprises 12 fully validated personas that cover \emph{minimal} to \emph{severe} depression and has already served as the ground-truth backbone of the eRisk 2025 pilot.



In summary, our contributions are fourfold:

(1) We propose the first open-source, BDI-II-based pipeline for generating clinically coherent simulated patients.

(2) We release a vetted repository of 12 depression personas together with conversation templates and logging scripts.\footnote{The full resource, code, and evaluation sheets are publicly available at: \url{https://github.com/Anxo06/TalkDep/}}

(3) We provide extensive automatic and expert validation, showing that both LLM judges and human clinicians reliably recognise the intended severity levels.

(4) We demonstrate the immediate utility of the resource by empowering the eRisk 2025 conversational track, establishing a reproducible benchmark for future research.

    
    
    

Together, these contributions create a scalable framework for advancing dialogue-based depression screening technology and studying how language models encode and express mental health symptomatology.

\section{Related Work on Patient Simulation}

Patient simulators have been widely used for educational purposes and the training of healthcare professionals \cite{good2003patient,bickley2012bates,papanagnou2021developing}. Typical examples include Standardised Patients (SPs), which stands for a method that employs people trained to act as patients and portray specific clinical cases \cite{barrows1993overview}. Recently, several studies have investigated the potential of using LLMs to support depression screening and even diagnosis \cite{sadeghi2023exploring,shin2024using,bao2024explainable}. However, most studies investigated the value of LLMs in addressing depression assessments. Only a few existing studies have explored patient simulation using LLMs. These include prompting LLMs with essential attributes (for example, situation, belief and emotion categories) \cite{wang2024patient}, or involving domain experts in a principle-based prompting pipeline \cite{louie2024roleplay}. Compared with existing efforts, this study contributes a resource that focusses on the availability of realistic user profiles and the advancement of prompt pipelines, which can be easily integrated with existing approaches.

\section{Methodology}

In this section, we describe the two major components of \texttt{TalkDep}, including the preparation of realistic patient profiles and the introduction of a simulator construction pipeline based on in-context learning (ICL). Figure \ref{fig:overview} presents an overview of the patient simulation process of \texttt{TalkDep}. 

\begin{table}[ht]
  \centering
  \small
  \caption{BDI-II scores grouped by depression level of the $12$ simulated LLM personas.}
  \label{tab:persona-severity}
  \begin{tabular}{@{}l l c | l l c@{}}
    \toprule
    \textbf{Level} & \textbf{Persona} & \textbf{BDI-II} &
    \textbf{Level} & \textbf{Persona} & \textbf{BDI-II} \\
    \midrule
    \multirow{3}{*}{Severe}   & Maria   & 40 & \multirow{3}{*}{Moderate} & Linda & 28 \\
                              & Marco   & 38 &                           & Laura & 23 \\
                              & Elena   & 35 &                           & James & 22 \\
    \midrule
    \multirow{3}{*}{Mild}     & Alex    & 15 & \multirow{3}{*}{Minimal}  & Priya &  7 \\
                              & Gabriel & 13 &                           & Maya  &  6 \\
                              & Ethan   & 12 &                           & Noah  &  5 \\
    \bottomrule
  \end{tabular}
\end{table}

\subsection{Patient Profile Construction}



For the construction of realistic patient profiles, we adopt a clinician-in-the-loop strategy. We employ a team of three clinical psychologists to collaboratively design a template that summarises biographical and clinical information to allow for an effective presentation of simulated patients. The evaluation of the corresponding templates is based on free-form interviews between clinicians and the corresponding personas to ensure that they support natural conversation and reflect their intended level of BDI-II and symptoms. The final template (left part of Figure~\ref{fig:overview}) comprises eight attribute groups, including the main ones\footnote{Due to page limitations, full attributes, prompt files, and materials are available in the project repository.}:

\begin{itemize}

    \item Persona information. Basic demographic details (name, age, gender, and a predefined BDI-II score). This demographic context and the specification of a BDI-II score in the persona shape the way depression manifests and is perceived~\cite{buckman2021role}.

    \item Key negative symptoms. Up to four key BDI-II symptoms that the LLM should manifest coherently. Real patients rarely exhibit all possible symptoms; instead, they present a few prominent ones, so ensuring the persona expresses some of these negative cognitions will reflect authentic depression experiences~\cite{tolentino2018dsm}. 
    
    \item Memory and reflection. Providing the simulated patient with a background (personal history and social context) is crucial for narrative coherence. Depression is frequently associated with adverse life events (for example, loss of a loved one, unemployment, trauma). In psychiatric interviewing, the acquisition of a social and life history is considered one of the most important pieces of the evaluation~\cite{smirnova2018language}. 

    \item Communication style. Linguistic markers can be strong indicators of depression~\cite{smirnova2018language}. The language of depressed patients often features a reflective focus on the past rather than the future. For example, frequent use of past-tense verbs and fewer future-orientated words~\cite{al2018absolute}. We define the persona’s vocabulary and sentence style to match these findings.
    
    
    


\end{itemize}
 
\subsection{In-Context Learning (ICL) Simulation Advancement}

Inspired by the recent success of in-context learning approaches in generating high-quality natural language conversational responses \cite{meade2023using,omidvar2023empowering}, we propose to further enrich patient profiles with example conversations that effectively reflect the severity of depression and key symptoms of major depression for each corresponding persona. As illustrated in Figure \ref{fig:overview}, we develop a multistep pipeline for the generation of synthetic dialogue.

Based on the patient profiles constructed in the initial stage, we begin with an initial prompt for generating \textbf{\textit{5}} synthetic dialogues (LLM No. 1 in Figure \ref{fig:overview}): one to represent the overall severity of depression and four others to highlight the individual main symptoms. The results of this first step are the synthetic conversations between the simulated patient and the LLM No. 1 instructed to act as an experienced therapist.

We then collect these generated dialogues, frame them as recorded counseling sessions between the patient and a therapist, and use another LLM (LLM No. 2 in Figure \ref{fig:overview}) to assess the severity of depression and the presence of major symptoms as a clinical professional. This results in a prediction of the severity of depression\footnote{The ideal pipeline would include a clinician-in-the-loop for this assessment, and we plan to incorporate this refinement in the next stage of our work.}.

By comparing the evaluation results with the ground truth depression profile, we assess whether conversations reflect the same level of depression severity, defined as an absolute difference in the BDI score of less than five points. If the evaluation results meet the requirement, we combine the generated conversations with the corresponding patient profile to form a complete context to initiate the final patient simulator (LLM No. 3 in Figure \ref{fig:overview}). Otherwise, we repeat the generation and evaluation process with a refined prompt for improved generation. As a result, by completing the simulation pipeline of \texttt{TalkDep}, we obtain a total of 12 simulated personas, listed in Table~\ref{tab:persona-severity}, representing various levels of severity of depression and a variety of major depression symptoms.

\section{Experimental Evaluation and Analysis}

\subsection{LLMs as Judges: A Pairwise Depression Comparison Experiment}

Since our simulated patient personas are designed around clinically BDI-II based depression levels (from 0 to 63 possibly score), we also evaluate whether external LLMs can accurately detect and differentiate these levels. To this end, we design a pairwise-ranking experiment in which a third-party LLM acts as an clinician and decides which of two conversation transcripts reflects a higher risk of depression. By testing how well the LLM judges align with the ground truth, we can gain insight into both the quality of our persona simulations and the sensitivity of LLMs to depression cues in dialogue. This experiment includes the following steps:

\textbf{(1) Conversation Generation:} As we explained in the previous section, we leverage semi-structured conversations with each simulated LLM persona in Table~\ref{tab:persona-severity}.
\textbf{(2) Pairwise Comparison:} Each transcript pair generated is given to the LLM judge, which identifies which of the two personas appears to have a higher risk of depression: Person A $>$ Person B, Person B $>$ Person A, or \textit{"Neither"} (insufficient evidence to make the comparison). For example, Maria has a higher BDI-II depression score than Priya, thus, Maria > Priya in the comparison.
\textbf{(3) Ranking Evaluation:} Finally, we compare these comparisons against the ground-truth BDI-II scores. We consider a correct prediction when the LLM acting as a judge chooses the most severe LLM persona from the comparison.

We conducted our experiments including four open-source models that vary in parameter count and training paradigm: llama3-8b-instruct \cite{grattafiori2024llama}, deepseek-r1-14b \cite{guo2025deepseek}, gemma-3-27b \cite{mesnard2024gemma} and llama3-70b-instruct \cite{grattafiori2024llama}. Since each persona is compared to each other in a pairwise approach, the experiment involves a total of 66 comparisons (\( \binom{12}{2} = 66 \)). The results, presented in Table~\ref{tab:pairwise-comparison-results}, highlight both the general precision of the prediction and key trends regarding incorrect predictions and ``Neither'' responses.

\begin{table}[t]
\centering
\caption{Pairwise comparison results for the LLM judges. 
Accuracy is computed as the percentage of correct comparisons. 
Error pairs are divided into those where the model confused personas with the same/different depression levels.}
\label{tab:pairwise-comparison-results}
\resizebox{.9\columnwidth}{!}{
\begin{tabular}{lccc}
\toprule
\textbf{Model} & \textbf{Accuracy} & \makecell[c]{\textbf{Error Pairs} \\ \textit{(same / different) level}} & \textbf{Neither} \\
\midrule
\emph{Llama3.1:8B}       & 75.81\% & 11.29\% / 12.95\% & 4 pairs \\
\emph{Deepseek-r1:14B}       & 86.36\% & 6.06\% / 7.58\% & 0 pairs \\
\emph{Gemma3.1:27B}   & 77.27\% & 12.12\% / 10.61\% & 0 pairs \\
\emph{Llama3.1:70B}       & 81.67\% & 8.33\% / 9.99\% & 6 pairs \\

\bottomrule
\end{tabular}}
\end{table}

\textbf{Overall performance.} Among the four judges, the DeepSeek R1-14B variant achieved the best accuracy (86.36\%) and never decided on the ``Neither'' option. Llama 3.1 70B ranked second (81.67\%) but declined to decide in six comparisons, indicating a more cautious policy. The two remaining models: Gemma 3.1 27B and Llama 3.1 8B obtained around 77\% and 76\% accuracy. These results show that model size alone does not guarantee higher accuracy, and that explicit uncertainty is only present in the Llama models.

\textbf{Incorrect Predictions.} All models shared certain challenging cases. For instance, \emph{Marco} (BDI-II=38, severe level) was misclassified a few times, whether compared to Moderate-level personas such as \emph{Linda} (BDI-II=28) or \emph{Laura} (BDI-II=23). This result could indicate that the Marco transcripts, though intended to convey severe symptoms, did not provide enough explicit cues for the model to confidently rank him above someone in the moderate category.

\textbf{Pair errors with the same vs. different depression levels.} Errors where the pairs come from the same level are less concerning from a diagnostic perspective, as individuals within the same BDI-II range (e.g., \emph{Alex} at 15 vs.\ \emph{Gabriel} at 13) can present highly similar clinical profiles. As Table \ref{tab:pairwise-comparison-results} shows, all LLMs make roughly the same number of mistakes when considering the same or different levels of depression. The errors in comparison pairs with different depression levels are low (no higher than 12.95\%), so even when the models are wrong, they rarely mix patients whose depression levels are clearly different.

\subsection{Clinical Professional Assessment for Simulation Validation}

To complement the automatic evaluation of LLM, we recruited two additional certified clinical psychologists (not involved in the simulation design) to rate the 12 LLM personas on a 1–5 Likert scale~\cite{joshi2015likert}. We asked them to fill out a form to examine the realism of simulated patients and their behaviours to reflect the severity of the encoded depression and key symptoms. The form covered two dimensions: ($i$) General interaction quality on three attributes: Humanness, Naturalness, Fluency. ($ii$) Depression diagnosis-oriented assessment, with four attributes: Emotional consistency, Symptom realism, Engagement/Responsiveness, Cognitive load \& processing style\footnote{The guidelines and evaluation instructions are accessible in the repository.}. Each rater conducted conversational interactions with each persona and received the corresponding full interview transcript to complete the form independently.  Table \ref{tab:persona_means} reports the mean scores (1-5) per LLM persona for all considered attributes.

\begin{table}[t]
\centering
\small
\setlength{\tabcolsep}{1.5pt} 
\caption{Mean attribute scores (1–5) for each LLM persona, grouped by depression-severity level.}
\label{tab:persona_means}
\resizebox{0.85\columnwidth}{!}{
\begin{tabular}{@{}l l c c c c c c c@{}}
\toprule
& & \multicolumn{3}{c}{\textbf{General interaction}} & \multicolumn{4}{c}{\textbf{Depression-oriented}} \\
\cmidrule(lr){3-5}\cmidrule(lr){6-9}
\textbf{Severity} & \textbf{Persona} & Hum. & Nat. & Flu. & Emo. & Sym. & Eng. & Cog. \\
\midrule
\multirow{3}{*}{Minimal} 
  & Maya  & 3.5 & 4.0 & 4.5 & 4.0 & 4.0 & 4.0 & 3.5 \\
  & Noah  & 3.0 & 3.0 & 4.0 & 3.5 & 4.0 & 3.5 & 4.0 \\
  & Priya & 3.5 & 4.5 & 4.5 & 4.0 & 4.5 & 4.0 & 4.5 \\
\midrule
\multirow{3}{*}{Mild} 
  & Alex    & 3.0 & 4.0 & 4.5 & 4.0 & 4.0 & 4.0 & 4.0 \\
  & Ethan   & 3.5 & 4.5 & 4.5 & 3.5 & 4.0 & 4.5 & 4.5 \\
  & Gabriel & 3.0 & 3.5 & 4.0 & 4.0 & 3.5 & 3.0 & 3.5 \\
\midrule
\multirow{3}{*}{Moderate} 
  & James & 3.5 & 4.0 & 4.5 & 3.5 & 4.5 & 4.0 & 3.5 \\
  & Laura & 4.0 & 4.5 & 4.5 & 4.5 & 4.5 & 3.5 & 4.5 \\
  & Linda & 3.5 & 4.0 & 4.5 & 3.5 & 3.5 & 3.5 & 3.0 \\
\midrule
\multirow{3}{*}{Severe} 
  & Elena & 4.5 & 4.5 & 4.5 & 4.5 & 4.0 & 3.5 & 3.5 \\
  & Marco & 4.0 & 4.0 & 4.5 & 4.0 & 4.5 & 3.5 & 3.5 \\
  & María & 4.0 & 4.0 & 4.5 & 3.5 & 4.0 & 3.0 & 3.0 \\
\bottomrule
\end{tabular}}
\end{table}

\textbf{Overall evaluation results.} Across all personas and attributes, the overall mean is 3.92, indicating that clinicians judged that the simulated patients were well above the midpoint of adequacy. Regarding the two dimensions, the general interaction attributes average 4.01, whereas the four depression-orientated items average 3.84. Thus, while language quality is perceived as strong, symptom portrayal remains slightly lower, suggesting that the pipeline captures psychological cues almost as convincingly as it renders conversational flow.

\looseness -1 \textbf{Differences based on severity levels.} Averaging scores within each severity level (Table \ref{tab:persona_means}) reveals little differences. The general quality of the interaction increases slightly with severity: from $3.83$ in the minimal group to $4.28$ in the severe group. Thus, clinicians perceived the more symptomatic personas equally or even more natural and fluent than the milder ones, suggesting that our prompt design maintained conversational coherence even when simulating severe cases. In contrast, the mean for depression-oriented attributes is better in the minimal and moderate bands ($3.96$ and $3.83$, respectively) and declines to $3.71$ for the severe band. This drop is primarily driven by lower scores of participation and cognitive load, which is consistent with the clinical expectation that severely depressed patients participate less and exhibit cognitive slowing~\cite{phillips2010implicit}.



\section{Findings, Discussions and Conclusions}

In general, our evaluation shows that the TalkDep simulation pipeline produces clinically relevant and conversationally coherent patient personas at a range of depression severity levels. Both LLMs and clinical experts as assessors confirmed that the simulated dialogues reliably reflect the intended BDI-II-based symptom profiles. LLM judges achieved up to 86\% precision in pairwise severity comparisons, indicating that the depression signals in our dialogues are detectable and interpretable. Clinical professionals rated personas highly on the dimensions related to interaction and symptoms, with an average score of 3.92 out of 5. People in the severe group were judged fluent and natural in conversations, while lower scores on engagement and cognitive activity matched clinical expectations. However, some difficulty remained in distinguishing adjacent severity levels, especially when moderate personas expressed strong emotional cues. This suggests a need for more controlled prompting to ensure a better representation of symptoms.

TalkDep has already proven useful on the eRisk 2025 pilot track and offers a clinically grounded benchmark to detect depression from conversations~\cite{parapar2025erisk}. Although current coverage is limited to 12 personas, future extensions could introduce more diverse profiles and modalities. In summary, our findings demonstrate that TalkDep offers a viable resource to benchmark dialogue-based depression detection systems. The evaluated personas also provide a solid foundation to support broader research on patient simulation, particularly for mental health education, where realistic and varied case presentations are essential for preparing clinical practitioners. These personas enables controlled experimentation with symptom severity, multi-system evaluation, and invites further research on the generation and assessment of simulated mental health dialogues. 

All stages of profile design and evaluation were performed in consultation with licenced clinicians and no real patient data was used. Because the patients are entirely synthetic, the resource poses no risk of re-identification or psychological harm to actual individuals.

\begin{acks}  
This work was supported by the project PID2022-137061OB-C21 (MCIN/AEI/10.13039/501100011033, Ministerio de Ciencia e Innovación, ERDF, \textit{A way of making Europe} by the European Union); the Consellería de Educación, Universidade e Formación Profesional, Spain (accreditations 2019-2022 ED431G/01 and GRC ED431C 2025/49); University of Alberta and University of Sheffield Seed Grant (X/016279-14); and the European Regional Development Fund, which supports the CITIC Research Center.  
\end{acks}

\section*{GenAI Usage Disclosure}
\label{sec:genai_usage_disclosure}
Generative AI tools were used only for minor editorial tasks such as proofreading, grammar and vocabulary refinement, and general language polishing during manuscript preparation. Every core idea, analysis, experiment, and section of prose was developed and written by the coauthors, with no AI-generated text included.



\bibliographystyle{ACM-Reference-Format}
\bibliography{sample-base}


\begin{thebibliography}{28}


\ifx \showCODEN    \undefined \def \showCODEN     #1{\unskip}     \fi
\ifx \showISBNx    \undefined \def \showISBNx     #1{\unskip}     \fi
\ifx \showISBNxiii \undefined \def \showISBNxiii  #1{\unskip}     \fi
\ifx \showISSN     \undefined \def \showISSN      #1{\unskip}     \fi
\ifx \showLCCN     \undefined \def \showLCCN      #1{\unskip}     \fi
\ifx \shownote     \undefined \def \shownote      #1{#1}          \fi
\ifx \showarticletitle \undefined \def \showarticletitle #1{#1}   \fi
\ifx \showURL      \undefined \def \showURL       {\relax}        \fi
\providecommand\bibfield[2]{#2}
\providecommand\bibinfo[2]{#2}
\providecommand\natexlab[1]{#1}
\providecommand\showeprint[2][]{arXiv:#2}

\bibitem[Al-Mosaiwi and Johnstone(2018)]%
        {al2018absolute}
\bibfield{author}{\bibinfo{person}{Mohammed Al-Mosaiwi} {and} \bibinfo{person}{Tom Johnstone}.} \bibinfo{year}{2018}\natexlab{}.
\newblock \showarticletitle{In an absolute state: Elevated use of absolutist words is a marker specific to anxiety, depression, and suicidal ideation}.
\newblock \bibinfo{journal}{\emph{Clinical psychological science}} \bibinfo{volume}{6}, \bibinfo{number}{4} (\bibinfo{year}{2018}), \bibinfo{pages}{529--542}.
\newblock


\bibitem[Bao et~al\mbox{.}(2024)]%
        {bao2024explainable}
\bibfield{author}{\bibinfo{person}{Eliseo Bao}, \bibinfo{person}{Anxo P{\'e}rez}, {and} \bibinfo{person}{Javier Parapar}.} \bibinfo{year}{2024}\natexlab{}.
\newblock \showarticletitle{Explainable depression symptom detection in social media}.
\newblock \bibinfo{journal}{\emph{Health Information Science and Systems}} \bibinfo{volume}{12}, \bibinfo{number}{1} (\bibinfo{year}{2024}), \bibinfo{pages}{47}.
\newblock


\bibitem[Barrows(1993)]%
        {barrows1993overview}
\bibfield{author}{\bibinfo{person}{Howard~S Barrows}.} \bibinfo{year}{1993}\natexlab{}.
\newblock \showarticletitle{An overview of the uses of standardized patients for teaching and evaluating clinical skills. AAMC}.
\newblock \bibinfo{journal}{\emph{Academic medicine}} \bibinfo{volume}{68}, \bibinfo{number}{6} (\bibinfo{year}{1993}), \bibinfo{pages}{443--51}.
\newblock


\bibitem[Bickley and Szilagyi(2012)]%
        {bickley2012bates}
\bibfield{author}{\bibinfo{person}{Lynn Bickley} {and} \bibinfo{person}{Peter~G Szilagyi}.} \bibinfo{year}{2012}\natexlab{}.
\newblock \bibinfo{booktitle}{\emph{Bates' guide to physical examination and history-taking}}.
\newblock \bibinfo{publisher}{Lippincott Williams \& Wilkins}.
\newblock


\bibitem[Buckman et~al\mbox{.}(2021)]%
        {buckman2021role}
\bibfield{author}{\bibinfo{person}{Joshua~EJ Buckman}, \bibinfo{person}{Rob Saunders}, \bibinfo{person}{Joshua Stott}, \bibinfo{person}{L-L Arundell}, \bibinfo{person}{Ciar{\'a}n O'Driscoll}, \bibinfo{person}{Molly~R Davies}, \bibinfo{person}{Thalia~C Eley}, \bibinfo{person}{Steven~D Hollon}, \bibinfo{person}{Tony Kendrick}, \bibinfo{person}{Gareth Ambler}, {et~al\mbox{.}}} \bibinfo{year}{2021}\natexlab{}.
\newblock \showarticletitle{Role of age, gender and marital status in prognosis for adults with depression: An individual patient data meta-analysis}.
\newblock \bibinfo{journal}{\emph{Epidemiology and psychiatric sciences}}  \bibinfo{volume}{30} (\bibinfo{year}{2021}), \bibinfo{pages}{e42}.
\newblock


\bibitem[Dozois et~al\mbox{.}(1998)]%
        {dozois1998psychometric}
\bibfield{author}{\bibinfo{person}{David~JA Dozois}, \bibinfo{person}{Keith~S Dobson}, {and} \bibinfo{person}{Jamie~L Ahnberg}.} \bibinfo{year}{1998}\natexlab{}.
\newblock \showarticletitle{A psychometric evaluation of the Beck Depression Inventory--II.}
\newblock \bibinfo{journal}{\emph{Psychological assessment}} \bibinfo{volume}{10}, \bibinfo{number}{2} (\bibinfo{year}{1998}), \bibinfo{pages}{83}.
\newblock


\bibitem[Good(2003)]%
        {good2003patient}
\bibfield{author}{\bibinfo{person}{ML Good}.} \bibinfo{year}{2003}\natexlab{}.
\newblock \showarticletitle{Patient simulation for training basic and advanced clinical skills}.
\newblock \bibinfo{journal}{\emph{Medical education}}  \bibinfo{volume}{37} (\bibinfo{year}{2003}), \bibinfo{pages}{14--21}.
\newblock


\bibitem[Grattafiori et~al\mbox{.}(2024)]%
        {grattafiori2024llama}
\bibfield{author}{\bibinfo{person}{Aaron Grattafiori}, \bibinfo{person}{Abhimanyu Dubey}, \bibinfo{person}{Abhinav Jauhri}, \bibinfo{person}{Abhinav Pandey}, \bibinfo{person}{Abhishek Kadian}, \bibinfo{person}{Ahmad Al-Dahle}, \bibinfo{person}{Aiesha Letman}, \bibinfo{person}{Akhil Mathur}, \bibinfo{person}{Alan Schelten}, \bibinfo{person}{Alex Vaughan}, {et~al\mbox{.}}} \bibinfo{year}{2024}\natexlab{}.
\newblock \showarticletitle{The llama 3 herd of models}.
\newblock \bibinfo{journal}{\emph{arXiv preprint arXiv:2407.21783}} (\bibinfo{year}{2024}).
\newblock


\bibitem[Guo et~al\mbox{.}(2025)]%
        {guo2025deepseek}
\bibfield{author}{\bibinfo{person}{Daya Guo}, \bibinfo{person}{Dejian Yang}, \bibinfo{person}{Haowei Zhang}, \bibinfo{person}{Junxiao Song}, \bibinfo{person}{Ruoyu Zhang}, \bibinfo{person}{Runxin Xu}, \bibinfo{person}{Qihao Zhu}, \bibinfo{person}{Shirong Ma}, \bibinfo{person}{Peiyi Wang}, \bibinfo{person}{Xiao Bi}, {et~al\mbox{.}}} \bibinfo{year}{2025}\natexlab{}.
\newblock \showarticletitle{Deepseek-r1: Incentivizing reasoning capability in llms via reinforcement learning}.
\newblock \bibinfo{journal}{\emph{arXiv preprint arXiv:2501.12948}} (\bibinfo{year}{2025}).
\newblock


\bibitem[Joshi et~al\mbox{.}(2015)]%
        {joshi2015likert}
\bibfield{author}{\bibinfo{person}{Ankur Joshi}, \bibinfo{person}{Saket Kale}, \bibinfo{person}{Satish Chandel}, {and} \bibinfo{person}{D~Kumar Pal}.} \bibinfo{year}{2015}\natexlab{}.
\newblock \showarticletitle{Likert scale: Explored and explained}.
\newblock \bibinfo{journal}{\emph{British journal of applied science \& technology}} \bibinfo{volume}{7}, \bibinfo{number}{4} (\bibinfo{year}{2015}), \bibinfo{pages}{396}.
\newblock


\bibitem[Lawrence et~al\mbox{.}(2024)]%
        {lawrence2024opportunities}
\bibfield{author}{\bibinfo{person}{Hannah~R Lawrence}, \bibinfo{person}{Renee~A Schneider}, \bibinfo{person}{Susan~B Rubin}, \bibinfo{person}{Maja~J Matari{\'c}}, \bibinfo{person}{Daniel~J McDuff}, {and} \bibinfo{person}{Megan~Jones Bell}.} \bibinfo{year}{2024}\natexlab{}.
\newblock \showarticletitle{The opportunities and risks of large language models in mental health}.
\newblock \bibinfo{journal}{\emph{JMIR Mental Health}} \bibinfo{volume}{11}, \bibinfo{number}{1} (\bibinfo{year}{2024}), \bibinfo{pages}{e59479}.
\newblock


\bibitem[Louie et~al\mbox{.}(2024)]%
        {louie2024roleplay}
\bibfield{author}{\bibinfo{person}{Ryan Louie}, \bibinfo{person}{Ananjan Nandi}, \bibinfo{person}{William Fang}, \bibinfo{person}{Cheng Chang}, \bibinfo{person}{Emma Brunskill}, {and} \bibinfo{person}{Diyi Yang}.} \bibinfo{year}{2024}\natexlab{}.
\newblock \showarticletitle{Roleplay-doh: Enabling Domain-Experts to Create LLM-simulated Patients via Eliciting and Adhering to Principles}. In \bibinfo{booktitle}{\emph{Proceedings of the 2024 Conference on Empirical Methods in Natural Language Processing}}. \bibinfo{pages}{10570--10603}.
\newblock


\bibitem[Lu et~al\mbox{.}(2024)]%
        {lu2024global}
\bibfield{author}{\bibinfo{person}{Bingqing Lu}, \bibinfo{person}{Lixia Lin}, {and} \bibinfo{person}{Xiaojuan Su}.} \bibinfo{year}{2024}\natexlab{}.
\newblock \showarticletitle{Global burden of depression or depressive symptoms in children and adolescents: A systematic review and meta-analysis}.
\newblock \bibinfo{journal}{\emph{Journal of affective disorders}} (\bibinfo{year}{2024}).
\newblock


\bibitem[Meade et~al\mbox{.}(2023)]%
        {meade2023using}
\bibfield{author}{\bibinfo{person}{Nicholas Meade}, \bibinfo{person}{Spandana Gella}, \bibinfo{person}{Devamanyu Hazarika}, \bibinfo{person}{Prakhar Gupta}, \bibinfo{person}{Di Jin}, \bibinfo{person}{Siva Reddy}, \bibinfo{person}{Yang Liu}, {and} \bibinfo{person}{Dilek Hakkani-Tur}.} \bibinfo{year}{2023}\natexlab{}.
\newblock \showarticletitle{Using In-Context Learning to Improve Dialogue Safety}. In \bibinfo{booktitle}{\emph{Findings of the Association for Computational Linguistics: EMNLP 2023}}. \bibinfo{pages}{11882--11910}.
\newblock


\bibitem[Mesnard et~al\mbox{.}(2024)]%
        {mesnard2024gemma}
\bibfield{author}{\bibinfo{person}{Thomas Mesnard}, \bibinfo{person}{Cassidy Hardin}, \bibinfo{person}{Robert Dadashi}, \bibinfo{person}{Surya Bhupatiraju}, \bibinfo{person}{Shreya Pathak}, \bibinfo{person}{Laurent Sifre}, \bibinfo{person}{Morgane Rivi{\`e}re}, \bibinfo{person}{Mihir~Sanjay Kale}, \bibinfo{person}{Juliette Love}, \bibinfo{person}{Pouya Tafti}, {et~al\mbox{.}}} \bibinfo{year}{2024}\natexlab{}.
\newblock \showarticletitle{Gemma: Open Models Based on Gemini Research and Technology}.
\newblock \bibinfo{journal}{\emph{CoRR}} (\bibinfo{year}{2024}).
\newblock


\bibitem[Moitra et~al\mbox{.}(2022)]%
        {moitra2022global}
\bibfield{author}{\bibinfo{person}{Modhurima Moitra}, \bibinfo{person}{Damian Santomauro}, \bibinfo{person}{Pamela~Y Collins}, \bibinfo{person}{Theo Vos}, \bibinfo{person}{Harvey Whiteford}, \bibinfo{person}{Shekhar Saxena}, {and} \bibinfo{person}{Alize~J Ferrari}.} \bibinfo{year}{2022}\natexlab{}.
\newblock \showarticletitle{The global gap in treatment coverage for major depressive disorder in 84 countries from 2000--2019: a systematic review and Bayesian meta-regression analysis}.
\newblock \bibinfo{journal}{\emph{PLoS medicine}} \bibinfo{volume}{19}, \bibinfo{number}{2} (\bibinfo{year}{2022}), \bibinfo{pages}{e1003901}.
\newblock


\bibitem[Omidvar and An(2023)]%
        {omidvar2023empowering}
\bibfield{author}{\bibinfo{person}{Amin Omidvar} {and} \bibinfo{person}{Aijun An}.} \bibinfo{year}{2023}\natexlab{}.
\newblock \showarticletitle{Empowering conversational agents using semantic in-context learning}. In \bibinfo{booktitle}{\emph{Proceedings of the 18th Workshop on Innovative use of NLP for Building Educational Applications (BEA 2023)}}. \bibinfo{pages}{766--771}.
\newblock


\bibitem[Papanagnou et~al\mbox{.}(2021)]%
        {papanagnou2021developing}
\bibfield{author}{\bibinfo{person}{Dimitrios Papanagnou}, \bibinfo{person}{Matthew~R Klein}, \bibinfo{person}{Xiao~Chi Zhang}, \bibinfo{person}{Kenzie~A Cameron}, \bibinfo{person}{Amanda Doty}, \bibinfo{person}{Danielle~M McCarthy}, \bibinfo{person}{Kristin~L Rising}, {and} \bibinfo{person}{David~H Salzman}.} \bibinfo{year}{2021}\natexlab{}.
\newblock \showarticletitle{Developing standardized patient-based cases for communication training: lessons learned from training residents to communicate diagnostic uncertainty}.
\newblock \bibinfo{journal}{\emph{Advances in Simulation}}  \bibinfo{volume}{6} (\bibinfo{year}{2021}), \bibinfo{pages}{1--11}.
\newblock


\bibitem[Parapar et~al\mbox{.}(2022)]%
        {erisk2022}
\bibfield{author}{\bibinfo{person}{Javier Parapar}, \bibinfo{person}{Patricia Mart{\'{\i}}n{-}Rodilla}, \bibinfo{person}{David~E. Losada}, {and} \bibinfo{person}{Fabio Crestani}.} \bibinfo{year}{2022}\natexlab{}.
\newblock \showarticletitle{Overview of eRisk 2022: Early Risk Prediction on the Internet}. In \bibinfo{booktitle}{\emph{Experimental {IR} Meets Multilinguality, Multimodality, and Interaction - 13th International Conference of the {CLEF} Association, {CLEF} 2022, Bologna, Italy, September 5–8, 2022}}. \bibinfo{pages}{233–256}.
\newblock


\bibitem[Parapar et~al\mbox{.}(2023)]%
        {erisk2023}
\bibfield{author}{\bibinfo{person}{Javier Parapar}, \bibinfo{person}{Patricia Mart{\'{\i}}n{-}Rodilla}, \bibinfo{person}{David~E. Losada}, {and} \bibinfo{person}{Fabio Crestani}.} \bibinfo{year}{2023}\natexlab{}.
\newblock \showarticletitle{Overview of eRisk 2023: Early Risk Prediction on the Internet}. In \bibinfo{booktitle}{\emph{Experimental {IR} Meets Multilinguality, Multimodality, and Interaction - 14th International Conference of the {CLEF} Association, {CLEF} 2023, Thessaloniki, Greece, September 18–21, 2023}}. \bibinfo{pages}{233–256}.
\newblock


\bibitem[Parapar et~al\mbox{.}(2024)]%
        {erisk2024}
\bibfield{author}{\bibinfo{person}{Javier Parapar}, \bibinfo{person}{Patricia Mart{\'{\i}}n{-}Rodilla}, \bibinfo{person}{David~E. Losada}, {and} \bibinfo{person}{Fabio Crestani}.} \bibinfo{year}{2024}\natexlab{}.
\newblock \showarticletitle{Overview of eRisk 2024: Early Risk Prediction on the Internet}. In \bibinfo{booktitle}{\emph{Experimental {IR} Meets Multilinguality, Multimodality, and Interaction - 15th International Conference of the {CLEF} Association, {CLEF} 2024, Grenoble, France, September 9-12, 2024, Proceedings, Part {II}}} \emph{(\bibinfo{series}{Lecture Notes in Computer Science}, Vol.~\bibinfo{volume}{14959})}, \bibfield{editor}{\bibinfo{person}{Lorraine Goeuriot}, \bibinfo{person}{Philippe Mulhem}, \bibinfo{person}{Georges Qu{\'{e}}not}, \bibinfo{person}{Didier Schwab}, \bibinfo{person}{Giorgio Maria~Di Nunzio}, \bibinfo{person}{Laure Soulier}, \bibinfo{person}{Petra Galusc{\'{a}}kov{\'{a}}}, \bibinfo{person}{Alba Garc{\'{\i}}a~Seco de~Herrera}, \bibinfo{person}{Guglielmo Faggioli}, {and} \bibinfo{person}{Nicola Ferro}} (Eds.). \bibinfo{publisher}{Springer}, \bibinfo{pages}{73--92}.
\newblock
\href{https://doi.org/10.1007/978-3-031-71908-0\_4}{doi:\nolinkurl{10.1007/978-3-031-71908-0\_4}}


\bibitem[Parapar et~al\mbox{.}(2025)]%
        {parapar2025erisk}
\bibfield{author}{\bibinfo{person}{Javier Parapar}, \bibinfo{person}{Anxo Perez}, \bibinfo{person}{Xi Wang}, {and} \bibinfo{person}{Fabio Crestani}.} \bibinfo{year}{2025}\natexlab{}.
\newblock \showarticletitle{eRisk 2025: contextual and conversational approaches for depression challenges}. In \bibinfo{booktitle}{\emph{European Conference on Information Retrieval}}. Springer, \bibinfo{pages}{416--424}.
\newblock


\bibitem[Phillips et~al\mbox{.}(2010)]%
        {phillips2010implicit}
\bibfield{author}{\bibinfo{person}{Wendy~J Phillips}, \bibinfo{person}{Donald~W Hine}, {and} \bibinfo{person}{Einar~B Thorsteinsson}.} \bibinfo{year}{2010}\natexlab{}.
\newblock \showarticletitle{Implicit cognition and depression: A meta-analysis}.
\newblock \bibinfo{journal}{\emph{Clinical Psychology Review}} \bibinfo{volume}{30}, \bibinfo{number}{6} (\bibinfo{year}{2010}), \bibinfo{pages}{691--709}.
\newblock


\bibitem[Sadeghi et~al\mbox{.}(2023)]%
        {sadeghi2023exploring}
\bibfield{author}{\bibinfo{person}{Misha Sadeghi}, \bibinfo{person}{Bernhard Egger}, \bibinfo{person}{Reza Agahi}, \bibinfo{person}{Robert Richer}, \bibinfo{person}{Klara Capito}, \bibinfo{person}{Lydia~Helene Rupp}, \bibinfo{person}{Lena Schindler-Gmelch}, \bibinfo{person}{Matthias Berking}, {and} \bibinfo{person}{Bjoern~M Eskofier}.} \bibinfo{year}{2023}\natexlab{}.
\newblock \showarticletitle{Exploring the capabilities of a language model-only approach for depression detection in text data}. In \bibinfo{booktitle}{\emph{2023 IEEE EMBS International Conference on Biomedical and Health Informatics (BHI)}}. \bibinfo{pages}{1--5}.
\newblock


\bibitem[Shin et~al\mbox{.}(2024)]%
        {shin2024using}
\bibfield{author}{\bibinfo{person}{Daun Shin}, \bibinfo{person}{Hyoseung Kim}, \bibinfo{person}{Seunghwan Lee}, \bibinfo{person}{Younhee Cho}, {and} \bibinfo{person}{Whanbo Jung}.} \bibinfo{year}{2024}\natexlab{}.
\newblock \showarticletitle{Using large language models to detect depression from user-generated diary text data as a novel approach in digital mental health screening: Instrument validation study}.
\newblock \bibinfo{journal}{\emph{Journal of Medical Internet Research}}  \bibinfo{volume}{26} (\bibinfo{year}{2024}), \bibinfo{pages}{e54617}.
\newblock


\bibitem[Smirnova et~al\mbox{.}(2018)]%
        {smirnova2018language}
\bibfield{author}{\bibinfo{person}{Daria Smirnova}, \bibinfo{person}{Paul Cumming}, \bibinfo{person}{Elena Sloeva}, \bibinfo{person}{Natalia Kuvshinova}, \bibinfo{person}{Dmitry Romanov}, {and} \bibinfo{person}{Gennadii Nosachev}.} \bibinfo{year}{2018}\natexlab{}.
\newblock \showarticletitle{Language patterns discriminate mild depression from normal sadness and euthymic state}.
\newblock \bibinfo{journal}{\emph{Frontiers in psychiatry}}  \bibinfo{volume}{9} (\bibinfo{year}{2018}), \bibinfo{pages}{105}.
\newblock


\bibitem[Tolentino and Schmidt(2018)]%
        {tolentino2018dsm}
\bibfield{author}{\bibinfo{person}{Julio~C Tolentino} {and} \bibinfo{person}{Sergio~L Schmidt}.} \bibinfo{year}{2018}\natexlab{}.
\newblock \showarticletitle{DSM-5 criteria and depression severity: implications for clinical practice}.
\newblock \bibinfo{journal}{\emph{Frontiers in psychiatry}}  \bibinfo{volume}{9} (\bibinfo{year}{2018}), \bibinfo{pages}{450}.
\newblock


\bibitem[Wang et~al\mbox{.}(2024)]%
        {wang2024patient}
\bibfield{author}{\bibinfo{person}{Ruiyi Wang}, \bibinfo{person}{Stephanie Milani}, \bibinfo{person}{Jamie Chiu}, \bibinfo{person}{Jiayin Zhi}, \bibinfo{person}{Shaun Eack}, \bibinfo{person}{Travis Labrum}, \bibinfo{person}{Samuel Murphy}, \bibinfo{person}{Nev Jones}, \bibinfo{person}{Kate Hardy}, \bibinfo{person}{Hong Shen}, {et~al\mbox{.}}} \bibinfo{year}{2024}\natexlab{}.
\newblock \showarticletitle{PATIENT-$\psi$: Using Large Language Models to Simulate Patients for Training Mental Health Professionals}. In \bibinfo{booktitle}{\emph{Proceedings of the 2024 Conference on Empirical Methods in Natural Language Processing}}. \bibinfo{pages}{12772--12797}.
\newblock


\end{thebibliography}


\end{document}